\documentclass{article}

     \PassOptionsToPackage{numbers, compress}{natbib}

\usepackage[preprint]{neurips_2025}

\usepackage[utf8]{inputenc} 
\usepackage[T1]{fontenc}    
\usepackage{hyperref}       
\usepackage{url}            
\usepackage{booktabs}       
\usepackage{amsfonts}       
\usepackage{nicefrac}       
\usepackage{microtype}      
\usepackage{xcolor}         

\usepackage{amsmath, amsthm}
\newtheorem{defination}{Defination}
\newtheorem*{hypothesis}{Editing Rules}
\usepackage{multirow}
\usepackage{tcolorbox}

\newcommand{\extend}[1]{\large{$\rightarrow$ #1}}

\usepackage{booktabs,caption,subcaption}
\usepackage{siunitx}
\usepackage{enumitem}
\usepackage{algorithm}
\usepackage{algorithmic}
\usepackage{xcolor}

\usepackage[bottom]{footmisc}  

\usepackage{graphicx} 
\usepackage{bm,dsfont}
\usepackage{soul}
\usepackage{tablefootnote}
\usepackage{subcaption}
\usepackage{wrapfig}
\usepackage{adjustbox}
\usepackage{booktabs} 

\hypersetup{
    colorlinks=true,
    linkcolor=blue,       
    citecolor=blue,        
    filecolor=magenta,      
    urlcolor=black          
}

\title{NeuralDB: Scaling Knowledge Editing in LLMs to 100,000 Facts with Neural KV Database}

\author{%
  Weizhi Fei \thanks{Department of Mathematical Sciences, Tsinghua University; \{fwz22, shih22, pjc22\}@mails.tsinghua.edu.cn}
  \And
  Hao Shi \textcolor{blue}{$^{*}$}
  \And
    Jing Xu \thanks{IIIS, Tsinghua University; \{xujing21, jingzhaoz\}@mails.tsinghua.edu.cn}
  \And
  Jingchen Peng \textcolor{blue}{$^{*}$}
  \And
  Jiazheng Li \thanks{College of AI, Tsinghua University; foreverlasting1202@outlook.com}
  \AND
    Jingzhao Zhang \textcolor{blue}{$^{\dagger}$} \textcolor{blue}{$^{\P}$}
  \And
  Bo Bai \thanks{Huawei Technologies Co., Ltd.; \{baibo8, harvey.hanwei, chenzhenyuan, niuxueyan3\}@huawei.com}
  \And
  Wei Han \textcolor{blue}{$^{\S}$}
  \And
  Zhenyuan Chen  \textcolor{blue}{$^{\S}$}
  \And
  Xueyan Niu \textcolor{blue}{$^{\S}$} \thanks{Corresponding authors.} 
}

\begin{document}

\maketitle

\begin{abstract}
Efficiently editing knowledge stored in large language models (LLMs) enables model updates without large-scale training. One possible solution is Locate-and-Edit (L\&E), allowing simultaneous modifications of a massive number of facts. However, such editing may compromise the general abilities of LLMs and even result in forgetting edited facts when scaling up to thousands of edits.  In this paper, we model existing linear L\&E methods as querying a Key-Value (KV) database. From this perspective, we then propose NeuralDB, an editing framework that explicitly represents the edited facts as a neural KV database equipped with a non-linear gated retrieval module,  
effectively preserving the general abilities of LLMs. 
Comprehensive experiments involving the editing of 10,000 facts were conducted on the ZsRE and CounterFacts datasets, using GPT2-XL, GPT-J (6B) and Llama-3 (8B). The results demonstrate that NeuralDB not only excels in editing efficacy, generalization, specificity, fluency, and consistency, but also preserves overall performance across six representative text understanding and generation tasks. Further experiments indicate that NeuralDB maintains its effectiveness even when scaled to 100,000 facts (\textbf{50x} more than in prior work).
\end{abstract}

\section{Introduction}

Updating the knowledge stored in the parameters of Large Language Models (LLMs) is crucial to refreshing outdated information \citep{zhu2024your} and integrating domain-specific knowledge to facilitate customization \citep{ge2023openagi}. However, retraining LLMs from scratch is often impractical due to the substantial computational resources and time required. Fine-tuning, while a more feasible approach, can lead to catastrophic forgetting \citep{luo2023empirical,gekhman2024does}. To address these challenges, \textit{knowledge editing} (KE) methods \citep{edit_survey2, MEND, IKE} have emerged as promising solutions that enable precise and cost-effective modifications of specific factual associations within LLMs.

Editing a large amount of knowledge without hurting LLMs~\citep{MEMIT} is an important but challenging task in KE. \textit{Locate-and-edit} (L\&E) methods~\citep{meng2022locating,PMET,fang2025alphaedit} are the main solutions to achieve this goal, and these methods are the first to accomplish efficient and scalable editing of massive facts. The L\&E paradigm aims to correct the activation of the models for new facts by introducing a linear perturbation $\Delta$ to the target parameter $\mathbf{W}$ within the feedforward network (FFN) layers~\citep{geva-etal-2021-transformer}. The new activations are learned from the new facts. 
Additionally, the activations of ``preserved knowledge'' that is unrelated to the edited content should be retained. Typically, the preserved knowledge is sampled from Wikipedia. The modified parameters $\Delta$ are then solved using least squares to ensure that the residuals required are generated for the new facts without affecting the samples to be preserved~\citep{meng2022locating}. Notably, AlphaEdit~\citep{fang2025alphaedit} introduces null space projection to enhance the preservation of the sampled knowledge, effectively maintaining the general capabilities of LLMs. Inspired by their work, we propose a unified framework for these methods, which enables us to further enhance the editing capacity.

\begin{figure}
    \vspace{-3em}
    \begin{minipage}{0.48\textwidth}
     \centering
     \includegraphics[width=\linewidth]{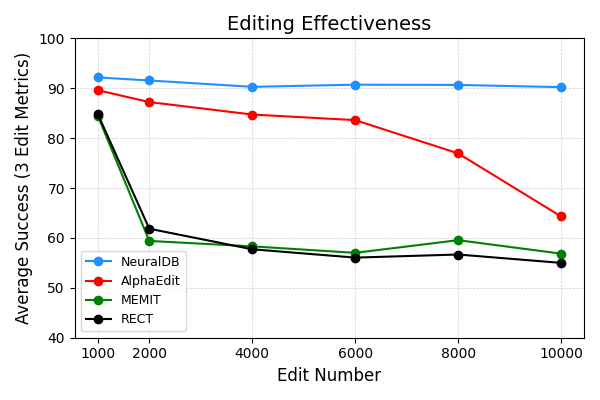}
   \end{minipage}\hfill
   \begin{minipage}{0.48\textwidth}
     \centering
     \includegraphics[width=\linewidth]{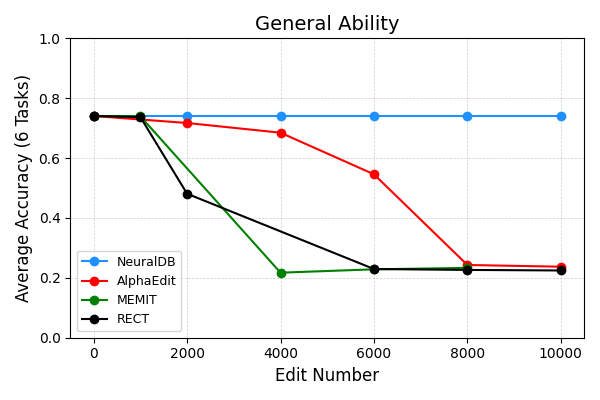}
   \end{minipage}
   \captionsetup{aboveskip=2pt, belowskip=10pt}
    \caption{
\textbf{The proposed NeuralDB scales the number of edited facts up to $10,000$ with almost no performance loss.} \textit{Left:} 
Average of efficacy, generalization, and specificity on the MCF dataset compared with AlphaEdit, MEMIT, and RECT. As the number of edits grows, NeuralDB experiences only a slight decline, whereas the other three methods undergo progressively larger decreases.
\textit{Right:}
Average performance on six tasks (MMLU, SciQ, Commonsense QA, ARC Challenge, Lambada, WSC273) from the lm-evaluation-harness benchmark. Preserving these general abilities is crucial, as they are acquired through large-scale pre-training on extensive datasets. NeuralDB \textbf{perfectly preserves} these general abilities even as the number of edited facts increases, whereas the other three methods lose them almost immediately as the number of edits grows.
}\label{fig:trade-off}
\vspace{-1em}
\end{figure}

Despite the progress, existing methods can only edit hundreds of facts without compromising the models' general abilities. When editing thousands of facts, the post-edited models usually suffer from a decline in the valuable general abilities developed from extensive training (see Fig.~\ref{fig:trade-off}).
The drop can be attributed to the sampled subset from Wikipedia being insufficient to represent general ability. Additionally, the previously updated information will fade as more facts are edited, because the linear system used by the editing methods is suboptimal in capacity.

In this paper, we demonstrate that most L\&E methods, including MEMIT~\citep{MEMIT}, D4S~\citep{huang2024reasons}, and AlphaEdit~\citep{fang2025alphaedit}, can be understood from the perspective of the Key-Value (KV) database. We conceptualize these methods as querying a KV database, wherein a certain hidden state serves as the query to retrieve the corresponding learned residual. Formally, the updated parameters of these methods can be interpreted as a weighted average of the residual matrix associated with the edited facts. We empirically investigate these weights across multiple post-edited models and show that they exhibit an extremely sparse form during inference. Specifically, when inferring on the edited fact, the weights closely resemble a one-hot vector, with only the weight corresponding to the edited fact being non-zero, thereby returning the associated residual. Conversely, when inference is made on unrelated content, the weights are in the form of a zero vector, thereby preventing interference.

In light of this new perspective, we propose a Neural KV Database (NeuralDB) editing framework, which integrates the target FFN layer with a gated non-linear retrieval module that replaces the original linear perturbation $\Delta$. NeuralDB first constructs the neural KV database from the edited facts and then utilizes the retrieval module to integrate this database into LLMs. The gated module returns the most compatible learned residual only when the post-edited models involve the edited facts, therefore, it overcomes the limitations of linearity in the L\&E methods and enjoys greater editing capacity. With the gated mechanism, our method can both protect the general ability, as well as reduce the computation and caching costs from sampling Wikipedia. Additionally, NeuralDB is easy to manage for supporting operations such as appending, modifying, and deleting.


To validate the capability of NeuralDB, we conducted comprehensive experiments on three models: GPT-2 XL~\citep{radford2019language}, GPT-J~(6B)~\citep{mesh-transformer-jax}, and Llama-3-Instruct~(8B)~\citep{grattafiori2024llama}.
These models were evaluated based on their performance for both edited facts and general capabilities. Our results demonstrate that NeuralDB achieves significantly better performance on edited facts, rephrased edited facts, and neighborhood facts, as well as preserves the fluency and consistency on the generation of post-edited models.
Moreover, NeuralDB exhibits great scalability --- the performances only drop $3\%$ as the edited facts scale from 2000 to 10,000 entries. 
To further validate the effectiveness of NeuralDB, we evaluated the post-edited models using widely adopted text understanding and generation tasks. The results indicate that NeuralDB can successfully edit tens of thousands of facts without compromising the quality of the generated text in Llama-3-8B. Extensive scaling experiments on 100,000 edited facts (45x more edited facts than AlphaEdit \cite{fang2025alphaedit}) demonstrate our method's ability to maintain high editing precision while fully preserving general language understanding and generation capabilities. This combination of large capacity without loss of generating quality underscores the potential of NeuralDB for trustworthy and customizable deployment of LLMs.

\section{Background}


Current KE methods typically focus on updating transformer-based LLMs with factual knowledge that can be represented as triples $(s,r,o)$, where $s$ denotes the subject, $r$ represents the relational predicate, and $o$ is the object. For instance, the fact ``The latest World Cup was held in Qatar.'' can be represented as $(\text{``The latest World Cup''}, \text{``was located in''}, \text{``Qatar''})$.
Conversely, triples can be transformed into natural language sentences, and we treat these two representations as interchangeable in the sequel. We denote the edited facts as a set of revised tuples $\mathcal{F}^* = \{(s_i,r_i,o_i \rightarrow  \hat{o}_i)\}$, where $\hat{o}_i$ represents the target new object that replaces the original $o_i$. Notably, KE should not compromise the general ability of the model~\citep{huang2024reasons}, as these abilities are usually developed from extensive pre-training.

As shown in Fig.~\ref{fig:framework}, during inference, the hidden state $\mathbf{h}^l$ of the prediction at the $l$-th FFN layer of a LLM is computed according to the following recursive form: 
\begin{equation}
 \mathbf{h}^l = \mathbf{h}^{l-1} + \mathbf{a}^{l} + \mathbf{m}^l, \quad \mathbf{m}^l = \mathbf{W}_{out}^l \sigma \left( \mathbf{W}_{in}^l(\mathcal{N}(\mathbf{h}^{l-1} + \mathbf{a}^{l})) \right),
\end{equation}  
where $\mathbf{a}^{l}$ and $\mathbf{m}^l$ are the outputs of the attention block and FFN layer, respectively, $\mathbf{W}_{in}^l$ and $\mathbf{W}_{out}^l$ represent the weight matrices of the $l$-th FFN layer, respectively, $\mathcal{N}(\cdot)$ represents the layer normalization, and $\sigma(\cdot)$ denotes the activation function. 
%
%
%
%
We follow previous research~\citep{meng2022locating,MEMIT} by modeling the FFN layer as operating linear key-value memories as follows:
\begin{equation}
    \mathbf{k}^l = \sigma \left( \mathbf{W}_{in}^l(\mathcal{N}(\mathbf{h}^{l-1} + \mathbf{a}^{l})) \right),\quad \mathbf{v}^l = \mathbf{W}_{out}^l \mathbf{k}^l.
\end{equation}
Then, the textual knowledge $(s,r,o)$ can be linked to the parametric knowledge of the LLM through the activation derived from the inference process. In this context, the key vector $\mathbf{k}$ can be interpreted as encoding the query $(s,r)$, while the object $o$ is subsequently decoded by the model based on the value vector $\mathbf{v}$ associated with the key~\citep{geva-etal-2021-transformer}. 

L\&E methods aim at adjusting the activation $\mathbf{v}$ on new facts by modifying the parameter. In this paradigm, a learnable perturbation is added to the activation $\mathbf{v}^l$ in the specified layer $l$ using supervised learning, resulting in a new activation $\hat{\mathbf{v}}^l$ that allows the model to generate new answers $\hat{o}$. 
Then the perturbation matrix $\Delta^l$ should satisfy 
\begin{equation}
    (\mathbf{W}_{out}^l +  \Delta^l) \mathbf{k}_i^l = \hat{\mathbf{v}}_i^l\quad \text{and}\quad (\mathbf{W}_{out}^l +  \Delta^l) \mathbf{k}_j^l =  \mathbf{v}_j^l
\end{equation}
where $\hat{\mathbf{v}}_i^l$ corresponds to the new facts and $(\mathbf{k}_j^l, \mathbf{v}_j^l)$ to the sampled knowledge that shall be preserved.


For simplified notation, we denote the parameter to be updated $\mathbf{W}_{out}^l \in \mathbb{R}^{d_2 \times d_1}$ by $\mathbf{W}$, where $d_1$ and $d_2$ represent the dimensions of the intermediate and output layers of the FFN, respectively. To update a large batch of facts, we obtain the key and value matrix by stacking the vectors: 
\begin{equation}\label{eq:KV}
    \mathbf{K}_1 = [\mathbf{k}_1,\mathbf{k}_2, \cdots, \mathbf{k}_m] \in \mathbb{R}^{d_1 \times m}, \quad\hat{\mathbf{V}}_1 = [\hat{\mathbf{v}}_1,\hat{\mathbf{v}}_2, \cdots, \hat{\mathbf{v}}_m] \in \mathbb{R}^{d_2 \times m},
\end{equation}
where $m$ is the number of edited facts, and we provide details on computing $\mathbf{k}_i$ and $\hat{\mathbf{v}}_i$ from the target editing facts in Appendix~\ref{App:kvr compute}. Additionally,  we define the residual matrix and residual vectors as 
\begin{equation}\label{eq:R}
\mathbf{R}_1 = \hat{\mathbf{V}}_1 - \mathbf{W} \mathbf{K}_1 \quad \text{and}\quad \mathbf{r}_i = \mathbf{R}_1[:, i].
\end{equation}
To preserve the general abilities of the edited model, current methods~\citep{MEMIT,huang2024reasons,fang2025alphaedit}, require the sampling of massive facts from Wikipedia\footnote{\url{https://huggingface.co/datasets/wikimedia/wikipedia}} to construct the matrix $\mathbf{K}_0$, which typically consists of 100,000 stacked key vectors representing preserved general knowledge.

\section{Rethinking locate-and-edit methods with querying Key-Value database}\label{Sec:Revisiting}

In this section, we demonstrate that most L\&E methods can be treated as querying a Key-Value (KV) database. We investigate these editing methods, which support massive knowledge updates, through both theoretical derivation and experimental validation. Specifically, we update all target facts in a single step in executing MEMIT~\citep{MEMIT} and AlphaEdit~\citep{fang2025alphaedit}, which has been shown to effectively mitigate the degradation of general capabilities by D4S~\citep{huang2024reasons}. For simplicity, we only discuss the parameter updating over a single FFN layer.

\begin{figure}
    \centering
    \includegraphics[width=1.0\linewidth]{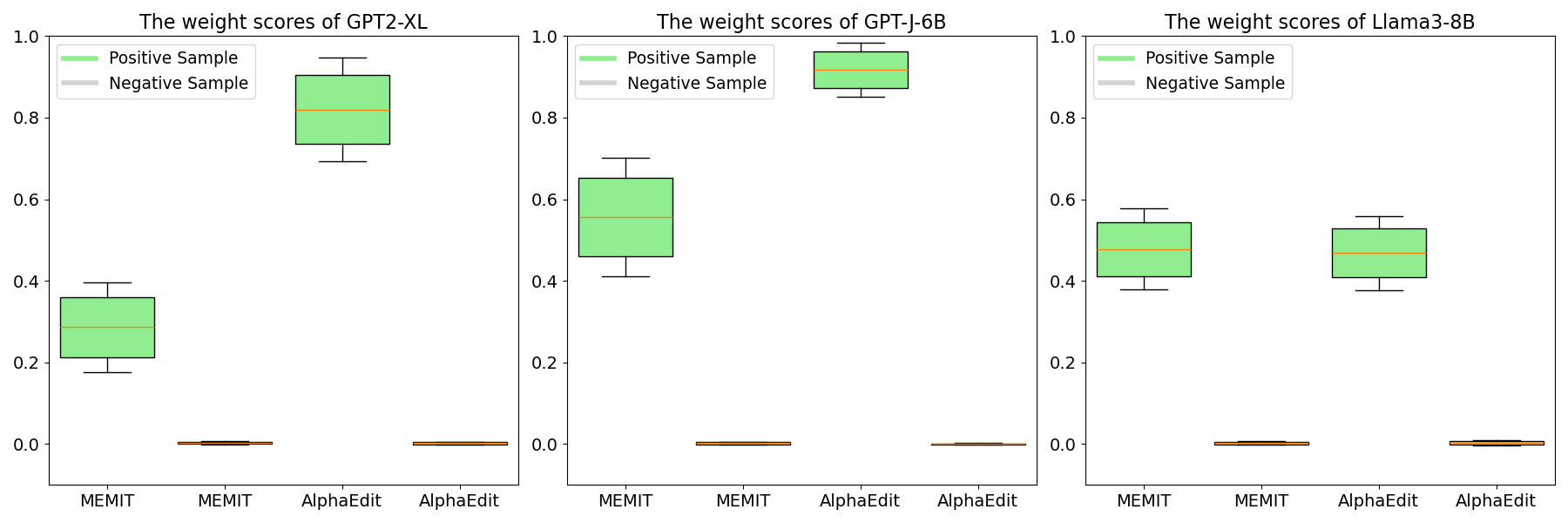}
    \caption{Visualization of weighted scores $\mathbf{\omega} = \mathbf{K}_1^T  \mathbf{S} \mathbf{k}$ using MEMIT and AlphaEdit for three models. The boxplots are generated from the mean and variance of weight scores, with the center line indicating the mean, boxes showing ±1 standard deviation, and whiskers ±1.5. When inferring the $i$-th edited fact, $\mathbf{\omega}_i$ serves as the positive sample, while the remaining elements of $\mathbf{\omega}$ are negative samples. The high positive score ensures that $\mathbf{R}_1 \mathbf{\omega}$ returns the right learned residual. }
    \label{fig:weighed scores}
    \vspace{-1em}
\end{figure}
We conclude that the operational mechanism of updating the parameter with $\Delta_{\text{upd}}$ can be written as
\begin{equation}
(\mathbf{W} + \Delta_{\text{upd}}) \mathbf{k} = \mathbf{v} + \mathbf{R}_1 \omega, \quad \omega =   \mathbf{K}_1^T \mathbf{S} \mathbf{k} \in \mathbb{R}^{m \times 1},
\end{equation}
where $\mathbf{k}$ and $\mathbf{v}$ are the key and value vectors from the original activation, $\mathbf{K}_1^T$ is the transpose of key matrix computed using edited facts, and $\mathbf{S}$ is the kernel matrix obtained from specific editing methods. $\mathbf{R}_1 \omega$ is the weighted average over $\mathbf{R}_1$, with weighted scores being self-similarity $\omega = \mathbf{K}_1^T \mathbf{S} \mathbf{k}$. This means that $\mathbf{k}$ serves as the query to retrieve information from the learned residuals $\mathbf{R}_1$. Then we derive the solutions for MEMIT and AlphaEdit, and present their structures.


The objective of MEMIT is the following constrained least squares optimization:
\begin{equation}
   \arg \min_{\Delta} \| (\mathbf{W} + \Delta ) \mathbf{K}_1 - \hat{\mathbf{V}}_1 \|_2^2 + \beta_1  \| \Delta \mathbf{K}_0 \|_2^2,
\end{equation}
where the term $\| \Delta \mathbf{K}_0 \|_2^2$ ensures that the updated parameters maintain preserved knowledge. 
The closed-form solution for MEMIT can be derived as follows:
\begin{equation}
\Delta_{\text{upd}}^{\text{MEMIT}} = \mathbf{R}_1 \mathbf{K}_1^T (\mathbf{K}_1 \mathbf{K}_1^T + \beta_1 \mathbf{K}_0 \mathbf{K}_0^T)^{-1}.
\end{equation}
Let $\mathbf{S}_1 =(\mathbf{K}_1 \mathbf{K}_1^T + \beta_1 \mathbf{K}_0 \mathbf{K}_0^T)^{-1}$ , then the update can be expressed as $\Delta_{\text{upd}}^{\text{MEMIT}} =  \mathbf{R}_1  \mathbf{K}_1^T  \mathbf{S}_1 $.



Similarly, we discuss AlphaEdit, which leverages the null space projection to convert soft constraints for preserved knowledge into hard constraints. The null space projection matrix $\mathbf{P}$ such that $\mathbf{P} \mathbf{K}_0 = \mathbf{0}$ is computed using SVD decomposition over $\mathbf{K}_0^T  \mathbf{K}_0$.  Then, AlphaEdit constructs $\Delta = \delta  \mathbf{P} $ such that  $\| \Delta  \mathbf{K}_0 \|$ approaches zeros and solves the following least squares problem with L2 Norm:
\begin{equation}
   \arg \min_{\delta} \| ( \mathbf{W} +  \mathbf{\delta}  \mathbf{P} )  \mathbf{K}_1 - \hat { \mathbf{V}}_1 \|_2^2 + \beta_2  \| \delta  \mathbf{P}   \|_2^2.
\end{equation}
The closed-form solution for AlphaEdit is 
\begin{equation}
\Delta_{\text{upd}}^{\text{AlphaEdit}} =    \mathbf{R}_1  \mathbf{K}_1^T  \mathbf{P}^T (  \mathbf{P}  \mathbf{K}_1  \mathbf{K}_1^T  \mathbf{P}^T + \beta_2  \mathbf{I})^{-1}  \mathbf{P}.
\end{equation}
Let $\mathbf{S}_2 = \mathbf{P}^T ( \mathbf{P} \mathbf{K}_1 \mathbf{K}_1^T\mathbf{P}^T + \beta_2 \mathbf{I})^{-1} \mathbf{P}$, this can be rewritten as $\Delta_{\text{upd}}^{\text{AlphaEdit}} = \mathbf{R}_1 \mathbf{K}_1^T \mathbf{S}_2$, which is also the weighted average over $\mathbf{R}_1$, with $\mathbf{K}_1^T \mathbf{S}_2 \mathbf{k}$ representing the weighted scores $\mathbf{\omega}$.  In particular, AlphaEdit returns $\mathbf{0}$ when $\mathbf{k}$ is from the null space of preserved knowledge $\mathbf{K}_0$. Consequently, $\mathbf{S}\mathbf{k} = \mathbf{P}^T ( \mathbf{P} \mathbf{K}_1 \mathbf{K}_1^T \mathbf{P}^T + \beta_2 \mathbf{I})^{-1} (\mathbf{P}\mathbf{k}) = \mathbf{0}$, which effectively preserves the general abilities.

We empirically visualize the weighted scores $\mathbf{\omega} = \mathbf{K}_1^T  \mathbf{S} \mathbf{k}$ for three post-edited models using MEMIT and AlphaEdit during their inference on new facts. Specifically, we perform KE on $1,000$ facts from the Counterfacts dataset and evaluate the post-edited models on these edited facts. When testing the $i$-th edited knowledge, we refer to the $i$-th component of $\mathbf{\omega}_i$ as the positive sample, while the remaining components are considered negative samples.  
As shown in Fig.~\ref{fig:weighed scores}, the weighted scores labeled as negative samples are close to $0$, while the weighted scores of positive samples are significantly higher, with those for AlphaEdit approaching $1$. These empirical results indicate that only the residuals corresponding to the edited facts are activated, while other residual vectors unrelated to the target facts are effectively suppressed. Additional results provided in the Appendix~\ref{subsec: weight score} demonstrate that current methods typically yield $\mathbf{\omega}=0$ when conducting inference on unmodified facts.


\section{Method}~\label{Sec: method}
From the perspective of querying KV database, we can improve the editing capacity and maintain the general abilities of LLMs by explicitly constructing the KV database. To this end, we propose a neural KV database (NeuralDB) editing framework, which serves as a plug-and-play module by replacing the linear perturbation with a non-linear gated retrieval module. Specifically, NeuralDB first computes the key vectors and residual vectors for the target editing facts and constructs a neural KV database. This constructed KV database is then equipped with the non-linear retrieval function for querying the appropriate residual vector during inference on the edited facts.

\begin{figure}[t]
  \centering
  \includegraphics[width=0.95\linewidth]{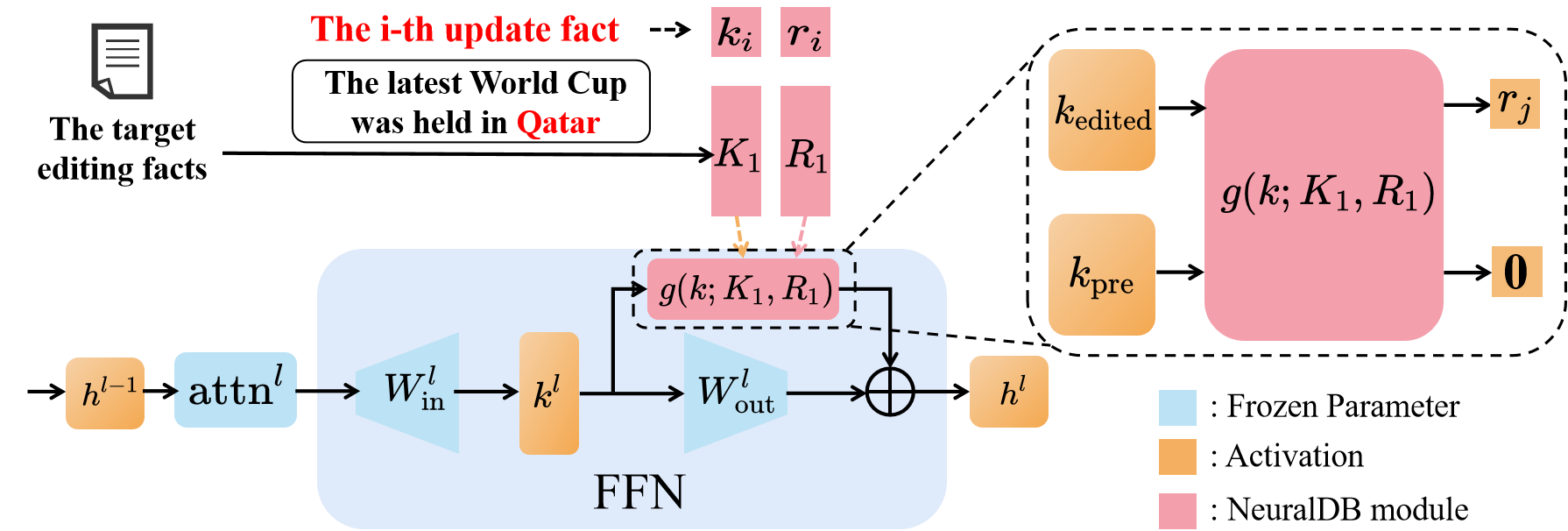}
  \caption{\textbf{Overview of NeuralDB editing framework.} We replace the linear system in L\&E with a non-linear gated retrieval module in the FFN layer. The key $\mathbf{K}_1$ and residual matrix $\mathbf{R}_1$, defined in Eq.~\ref{eq:KV} and Eq.~\ref{eq:R}, are computed to construct the neural KV database. During inference, our non-linear gated function $g(\cdot;\mathbf{K}_1,\mathbf{R}_1)$ only returns the most matched residual $\mathbf{r}_j$ when post-edited models infer one edited fact and involve key vectors $\mathbf{k}_{\text{edited}}$. The function $g(\cdot;\mathbf{K}_1,\mathbf{R}_1)$ reverts zero vector $\mathbf{0}$ when involving the key vector $\mathbf{k}_{\text{pre}}$ of preserved knowledge.}
  \vspace{-1em}
     \captionsetup{aboveskip=1pt, belowskip=-16pt}
  \label{fig:framework}
\end{figure}

\subsection{Neural KV database}

Given the target facts $\mathcal{F}^* = \{(s_i,r_i,o_i \rightarrow  \hat{o}_i)\}$, we first compute their key and residual matrix to obtain $\mathbf{K}_1$ and $\mathbf{R}_1$. Below, we formally define them as a neural KV database.
\begin{defination}[Neural Key-Value Database]
    Given $\mathcal{F}^*$, the constructed neural KV database can be represented as $(\mathbf{K}_1,\mathbf{R}_1)$, where $\mathbf{K}_1 \in \mathbb{R}^{d_1 \times m}$ and $\mathbf{R}_1 \in \mathbb{R}^{d_2 \times m}$ denote the key matrix and residual matrix of the edited facts as defined in Eq.~\eqref{eq:KV} and Eq.~\eqref{eq:R}. $\mathbf{K}_1$ and $\mathbf{R}_1$ serve as keys and values within the database, with $\mathbf{k}_i = \mathbf{K}_1[:,i]$ being associated with $\mathbf{r}_i = \mathbf{R}_1[:,i]$.
\end{defination}
%
%

We propose the following sufficient conditions for effective editing.

\begin{hypothesis}
Given a neural KV database $(\mathbf{K}_1, \mathbf{R}_1)$ for target facts $\mathcal{F}^*$. Let $\mathbf{k}$ denote the key vector obtained through inference on a sequence $\mathbf{x}$, 
KE should then adhere to the following rules:
\begin{enumerate}[label=\Roman*.]
    \item If the sequence $\mathbf{x}$ is related to the $i$-th edited fact, then $\mathbf{k}$ matches with $\mathbf{k}_j$ within $\mathbf{K}_1$, and the corresponding residual vector $\mathbf{r}_j$ is returned;
    \item If the sequence $\mathbf{x}$ is not related to any edited facts, then $\mathbf{k}$ does not match any keys within $\mathbf{K}_1$, and the zero vector $\mathbf{0}$ is returned.
\end{enumerate}
\end{hypothesis}
Rule I ensures the effective updating of target facts, while Rule II ensures the preservation of knowledge unrelated to the edited facts. To more strictly realize these rules, we propose the following non-linear retrieval function.

\subsection{Nonlinear gated retrieval module}

Given a neural KV database $(\mathbf{K}_1,\mathbf{R}_1)$ constructed from target edited facts, we propose the following function, which returns the residual with maximal similarity to involved the target edited facts:
\begin{equation}
    g(\mathbf{k};\mathbf{K}_1, \mathbf{R}_1) = \mathbf{r}_j * \overbrace{\mathbf{1}_{\cos(\mathbf{k},\mathbf{k}_j) > \gamma}}^{\text{Gate}} , j = \arg \max \cos(\mathbf{k},\mathbf{k}_i), 
\end{equation}
where $\cos(\cdot,\cdot)$ denotes the cosine similarity, $\mathbf{1}$ represents the indicator function, and $\gamma$ is the parameter controlling the gating mechanism. We employ cosine similarity because it  provides good interpretability, facilitating the easy setting of $\gamma$.

As illustrated in Fig.~\ref{fig:framework}, the nonlinear function is integrated into the target FFN layer as follows:
\begin{equation}
    \mathbf{v}^l = \mathbf{W}^l \mathbf{k}^l + g(\mathbf{k}^l; \mathbf{K}_1, \mathbf{R}_1).
\end{equation}
When post-edited models utilize preserved knowledge, the involved key vectors $\mathbf{k}$ typically exhibit low similarity with all key vectors $\mathbf{k}_i$ in the matrix $\mathbf{K}_1$. Therefore, this additional module will not provide the perturbation, ensuring that the original model's workflow remains same. Therefore, the general ability of pre-edited LLMs can successfully be preserved. In contrast, when post-edited models encounter revised facts, our retrieval function can identify and recall the most closely related residual vector, because of the high similarity between the keys. This precise retrieval mechanism effectively updates the specified parametric knowledge.


Another significant advantage of our proposed framework is its ease of deployment. Current L\&E methods typically update parameters through optimization, which poses challenges for incremental updates or reverting modified facts. In contrast, our framework maintains the neural KV database that corresponds specifically to the target facts. This design facilitates addition, deletion, and modification of edited facts within the model, thereby improving the flexibility of the editing process.

\paragraph{Single layer versus multi-layer editing} 
We just implement our module in one single FFN layer. Although previous L\&E methods typically employ multi-layer strategies, we find this strategy offers limited performance gains. We provide a related discussion in Appendix~\ref{app:abalation study}.

\paragraph{Additional memory usage and computation time} 
The introduction of additional parameters in the NeuralDB module incurs increased computation and storage requirements. Specifically, the space complexity is $O((d_1+d_2) \times m)$ for the storage of matrices $\mathbf{K}_1 \in \mathbb{R}^{d_1 \times m}$ and $\mathbf{R_1} \in \mathbb{R}^{d_2 \times m}$, where $m$ denotes the number of facts. When editing 10,000 facts with Llama 3 8B (Instruct), the additional parameter size amounts to 150M, which constitutes approximately $2.2\%$ of the original model's size. The evaluation time for $10,000$ facts in Counterfact dataset increases only by $1.5\%$ compared to the original model. Detailed information regarding additional memory usage and computation time across various models is provided in Appendix~\ref{App: addional memory and time}.

\section{Experiments}
In this section, we present a comprehensive evaluation of our method for mass KE. For fair comparison, we mostly follow the experimental setups in AlphaEdit~\citep{fang2025alphaedit} to benchmark our method. Specifically, we examine whether the edited models can effectively balance mastery of the edited facts with retention of their general capabilities. Detailed implementation of the NeuralDB editing framework is provided in Appendix~\ref{App:implementation details}. Additionally, ablation studies are included in Appendix~\ref{app:abalation study}, where we explore the impact of layer selection, the setting of the parameter $\gamma$, and the multi-layer configurations.

\paragraph{Models and methods} 
We select three representative LLMs, including GPT-2 XL~\citep{radford2019language}, GPT-J (6B)~\citep{mesh-transformer-jax}, and Llama3 (8B)~\citep{grattafiori2024llama}. We compare our method with the following KE methods, including Fine-Tune (FT)~\citep{MEMIT}, MEND~\citep{MEND}, ROME~\citep{meng2022locating}, MEMIT~\citep{MEMIT},  MELO~\citep{MELO}, RECT~\citep{gu2025editing}, and AlphaEdit~\citep{fang2025alphaedit}. We provide a detailed introduction on these baselines and models in Appendix~\ref{App:baselines}. 

\paragraph{Datasets for knowledge editing} 

To evaluate KE methods, we utilize two widely recognized benchmarks: the Counterfact dataset~\citep{meng2022locating} and the ZsRE dataset~\citep{zsre}. Consistent with previous research~\citep{meng2022locating,fang2025alphaedit}, we employ the following evaluation metrics: \textit{efficacy} (success of edited facts), \textit{generalization} (success of paraphrased facts), \textit{specificity} (success of neighboring facts), \textit{fluency} (generation entropy), and \textit{consistency} (reference score). Detailed explanations of the datasets and metrics are provided in Appendix~\ref{App:datasets and metrics}.

\paragraph{Datasets for general ability} 

We assess the general capabilities of the edited LLMs using the following typical datasets, SciQ~\citep{welbl-etal-2017-crowdsourcing} (science question answering), MMLU~\citep{hendrycks2021measuring} (massive multitask language understanding),  Commonsense QA~\citep{talmor-etal-2019-commonsenseqa} (commonsense knowledge understanding), ARC Challenge~\citep{clark2018think} (challenge task requiring reasoning), WSC273~\citep{kocijan-etal-2019-surprisingly} (coreference resolution), Lambada~\citep{paperno-etal-2016-lambada}(predict the endings of text passages). Datasets such as SciQ, MMLU, and Commonsense QA 
primarily evaluate knowledge-based question answering, focusing on the models' ability to understand and retain factual information. In contrast, the ARC Challenge, WSC273, and Lambada are designed to assess capabilities beyond mere knowledge memory, such as reasoning and text generation. Detailed information regarding these datasets is provided in Appendix~\ref{App:datasets and metrics}.

\subsection{The editing effectiveness of post-edited models}

\begin{table}[t]
\centering
\caption{Comparison of NeuralDB with existing methods on the massive KE task. We evaluated the performance of editing 2,000 facts, with results for Pre-edited, FT, MEND, InstructEdit, MELO, and ROME sourced from \citet{fang2025alphaedit}. For 10,000 facts, the results for MEMIT, RECT, AlphaEdit, and NeuralDB are denoted using the arrow $(\rightarrow)$ notation. The best results for both 2,000 and 10,000 facts are highlighted in bold, respectively.}
\vskip 0.05in
\large
\renewcommand{\arraystretch}{1.2}
\resizebox{\textwidth}{!}{
\begin{tabular}{cccccccccc}
\toprule[1.5pt]
\raisebox{-1.5ex}{\textbf{Method}} & \raisebox{-1.5ex}{\textbf{Model}}  & \multicolumn{5}{c}{\textbf{Counterfact}} & \multicolumn{3}{c}{\textbf{ZsRE}} \\
\cmidrule(lr){3-7} \cmidrule(lr){8-10}
&& \textbf{Efficacy$\uparrow$} & \textbf{Generalization$\uparrow$} & \textbf{Specificity$\uparrow$} & \textbf{Fluency$\uparrow$} & \textbf{Consistency$\uparrow$} & \textbf{Efficacy$\uparrow$} & \textbf{Generalization$\uparrow$} & \textbf{Specificity$\uparrow$} \\
\midrule[1pt]
Pre-edited & \multirow{9}{*}{\rotatebox{90}{{LLaMA3 }}}& {7.9} & {10.6} & {89.5} & {635.2} & {24.1} & {37.0} & {36.3} & {31.9}\\
\midrule
FT&  & 83.3 & 67.8 & 46.6 & 233.7 & {8.8} & {30.5} & {30.2} & {15.5}\\
MEND& & {63.2} & {61.2} & {45.4} & {372.2} & {4.2} & {0.9} & {1.1} & {0.5}\\
{InstructEdit}& & {{66.6}} & {{64.2}} & {{47.1}} & {{443.9}} & {{7.3}} & {{1.6}} & {{1.4}} & {{1.0}}\\
MELO&& {65.3} & {58.6} & {63.4} & {609.0} & {32.4} & {25.2} & {24.1} & {30.4}\\
ROME&& {64.4} & {61.4} & {49.4} & {449.1} & {3.3} & {2.0} & {1.8} & {0.7}\\
MEMIT & & {63.5\extend{63.4}} & {62.8\extend{56.6}} & {52.0\extend{50.55}} & {466.6\extend{460.4}} & {6.5\extend{6.5}} & {36.7\extend{0.1}} & {32.9\extend{0.1}} & {19.1\extend{1.5}} \\
RECT& & {64.2\extend{60.0}} & {62.5\extend{53.9}} & {58.9\extend{51.2}} & {502.8\extend{399.1}} & {12.9\extend{1.6}} & {86.8\extend{0.0}} & {82.3\extend{0.0}} & {31.9\extend{0.0}} \\
AlphaEdit  & & {99.1\extend{75.8}} & \textbf{94.0}\extend{63.1} & {68.6\extend{54.0}} & {622.7\extend{417.8}} & {32.8\extend{7.0}} & {94.4\extend{90.5}} & {91.3\extend{85.9}} & \textbf{32.6}\extend{30.3}\\
NeuralDB & & \textbf{99.9}\extend{\textbf{99.2}} & {86.6}\extend{\textbf{85.9}} & \textbf{88.2\extend{85.6}} & \textbf{632.7\extend{631.02}} & \textbf{32.9\extend{32.6}} & \textbf{96.3\extend{95.9}} & \textbf{92.0\extend{91.0}} & {31.9}\extend{\textbf{31.8}}\\
\midrule[1pt]
Pre-edited & \multirow{9}{*}{\rotatebox{90}{{GPT-J }}} & {16.2} & {18.6} & {83.1} & {621.8} & {29.7} & {26.3} & {25.8} & {27.4}\\
\midrule
FT&  & {92.2} & {72.4} & {43.4} & {297.9} & {6.7} & {72.4} & {68.9} & {19.7}\\
MEND& & {46.2} & {46.2} & {53.9} & {242.4} & {3.9} & {0.7} & {0.7} & {0.5}\\
{InstructEdit}& & {{50.6}} & {{51.7}} & {{56.3}} & {{245.9}} & {{4.2}} & {{0.9}} & {{0.9}} & {{0.7}}\\
MELO&& {78.3} & {60.5} & {66.8} & {610.8} & {24.3} & {82.2} & {32.9} & {26.7}\\
ROME& & {57.5} & {54.2} & {52.1} & {589.4} & {3.2} & {56.4} & {54.7} & {9.9}\\
MEMIT& & {98.6\extend{48.8}} & {95.4\extend{49.3}} & {66.1\extend{51.9}} & {557.8\extend{281.5}} & {36.5\extend{5.1}} & {90.5\extend{0.2}} & {84.7\extend{0.1}}   & \textbf{30.9}{\extend{0.2}}\\
RECT& & {98.8\extend{76.3}} & {86.3\extend{70.6}} & {74.4\extend{54.9}} & {618.1\extend{517.3}} & {41.2\extend{25.4}} & {96.6\extend{53.5}} & {91.5\extend{49.6}} & {29.0\extend{21.9}} \\
AlphaEdit & & \textbf{99.8}\extend{91.6} & \textbf{96.3}\extend{79.6} & {76.2\extend{60.3}} & {618.5\extend{517.8}} & \textbf{41.9}{\extend{6.9}} & {99.7\extend{94.2}} & \textbf{95.9}\extend{86.1} & {28.8\extend{22.5}}\\
NeuralDB & & {99.7}\extend{\textbf{99.1}} & {94.6}\extend{\textbf{93.2}} & \textbf{80.0\extend{75.7}} & \textbf{619.8\extend{620.0}} & {41.4}\extend{\textbf{41.3}} & \textbf{99.2\extend{98.2}} & \textbf{95.9\extend{95.0}} & {27.5}\extend{\textbf{27.0}}\\
\midrule[1pt]
Pre-edited & \multirow{9}{*}{\rotatebox{90}{{GPT2-XL }}} & {22.2} & {24.3} & {78.5} & {626.6} & {31.9} & {22.2} & {31.3} & {24.2}\\ \midrule
FT&  & {63.6} & {42.2} & {57.1} & {519.4} & {10.6} & {37.1} & {33.3} & {10.4}\\
MEND& & {50.8} & {50.8} & {49.2} & {407.2} & {1.0} & {0.0} & {0.0} & {0.0}\\
{InstructEdit}& & {{55.3}} & {{53.6}} & {{53.3}} & {{412.6}} & {{1.1}} & {{3.5}} & {{4.3}} & {{3.2}}\\
ROME& & {54.6} & {51.2} & {52.7} & {366.1} & {0.7} & {47.5} & {43.6} & {14.3}\\
MELO&& {72.6} & {53.6} & {63.3} & {588.6} & {23.6} & {93.5} & {45.3} & {23.5}\\
MEMIT& & {93.0\extend{58.5}} & {83.3\extend{55.8}} & {58.9\extend{56.1}} & {481.8\extend{496.2}} & {23.2\extend{8.1}} & {74.4\extend{3.5}} & {66.9\extend{2.8}} & \textbf{25.87}{\extend{2.07}}\\
RECT& & {91.8\extend{86.9}} & {79.5\extend{69.5}} & {64.0\extend{55.0}} & {482.1\extend{517.8}} & {20.3\extend{10.9}} & {82.6\extend{27.5}} & {74.7\extend{25.1}} & {24.6\extend{13.1}} \\
AlphaEdit & & {99.4\extend{92.2}} & {93.8\extend{76.5}} & {65.6\extend{56.5}} & {584.0\extend{580.9}} & {37.9\extend{29.6}} & {93.2\extend{57.1}} & {83.5\extend{47.5}} & {25.3\extend{13.5}} \\
NeuralDB & & \textbf{99.8\extend{99.1}} & \textbf{97.2\extend{95.7}} & \textbf{74.1\extend{70.9}} & \textbf{621.5\extend{619.9}} & \textbf{42.2\extend{41.7}} & \textbf{96.3\extend{94.6}} & \textbf{92.8\extend{91.0}} & {25.0}\extend{\textbf{24.2}} \\
\bottomrule[1.5pt]
\end{tabular}
}
\vspace{-1em}

\label{tab:overall_comp}
\end{table}

We evaluated the performance after editing 2,000 facts and also included the results of 10,000 facts in parentheses for MEMIT, RECT, AlphaEdit, and NeuralDB. The results are presented in Table~\ref{tab:overall_comp}. Our method achieves superior performance on all metrics in three models and two datasets. In particular, when scaling the number of edited facts from 2,000 to 10,000, our method exhibits only a negligible performance drop, in contrast to the substantial degradation observed in the baselines.

Our method demonstrates exceptional efficacy across various scenarios. Even after editing 10,000 facts, our post-edited models maintain strong efficacy and specificity. In contrast, all baseline methods exhibit a significant decline in both efficacy and specificity metrics. This highlights our method's excellent scalability in effective editing.

In terms of specificity and fluency metrics, our post-edited models demonstrate nearly identical effectiveness to the pre-edited models, representing a substantial improvement over the baselines on the CounterFacts dataset. Furthermore, our method ensures that the generated text maintains coherence with a high degree of consistency. This indicates that our method can preserve the memory of unmodified facts while maintaining its generative capabilities.

\subsection{The general abilities of post-edited models}

\begin{figure}[t]
    \centering
    \includegraphics[width=1.0\linewidth]{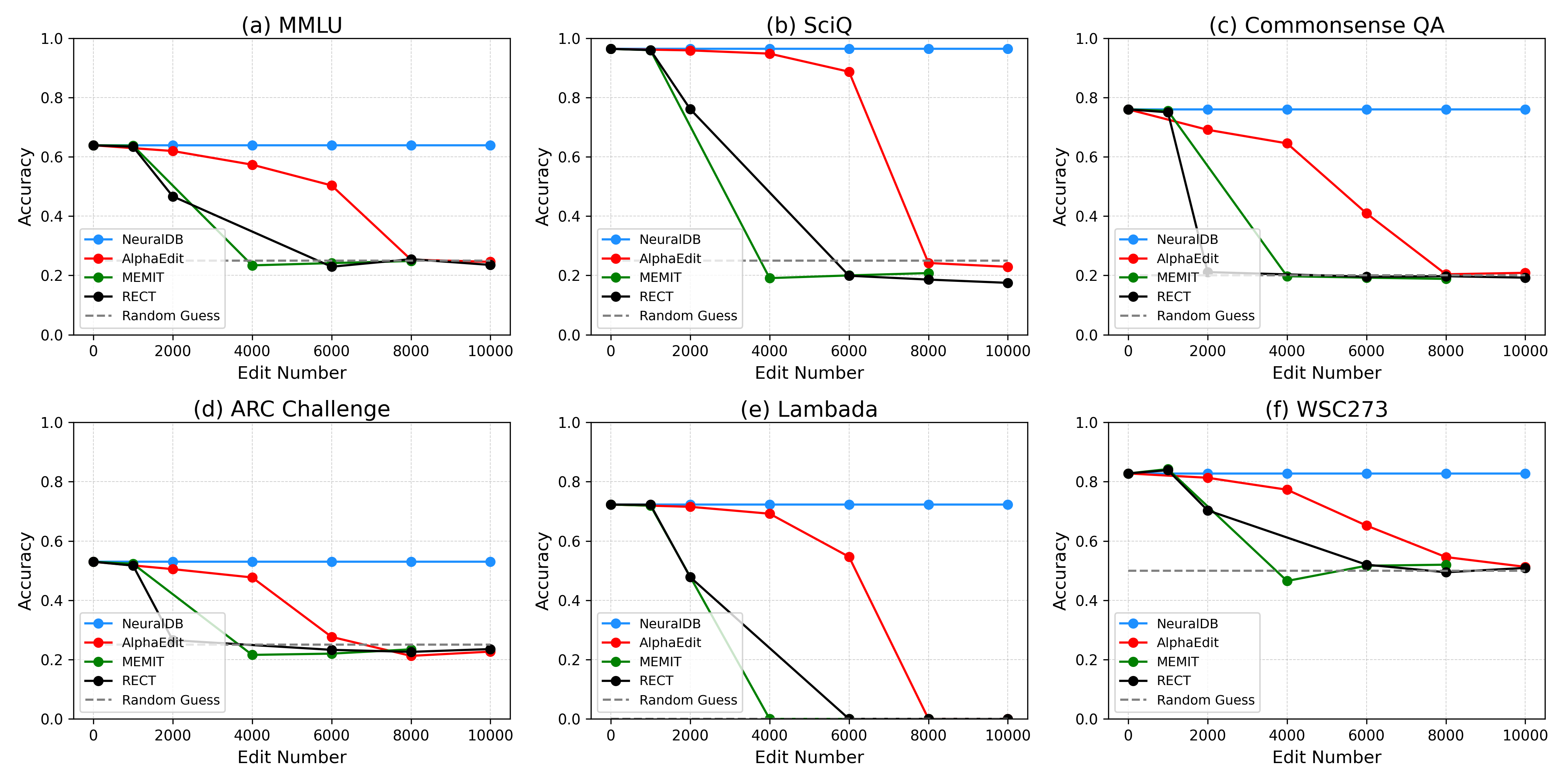}
    \vspace{-2em}
    \caption{Results of general abilities after massive editing. The performance of NeuralDB is compared with baseline methods, including MEMIT, RECT, and AlphaEdit. The random guessing baseline for multi-choice datasets is indicated by dashed lines. NeuralDB demonstrates exceptional preservation of general abilities when the number of edits scales up.}
    \captionsetup{aboveskip=-1pt, belowskip=-12pt}
    \label{fig:lmbench}
       \vspace{-1em}

\end{figure}

We assessed post-edited models with various configurations, evaluating their performance across 2,000, 4,000, 6,000, 8,000, and 10,000 facts, as depicted in Fig.~\ref{fig:lmbench}. The evaluation was conducted on lm-evaluation-harness~\citep{eval-harness}. 
%
%
%
The results show that our method effectively edits a large number of facts without compromising the general abilities of the models across various tasks. In contrast, existing L\&E methods struggle with 4,000 facts editing and exhibit a rapid decline in general abilities as the number of edited facts increases. Notably, these baseline methods achieve favorable results on the SciQ dataset, likely due to the dataset's content being well-represented in Wikipedia and thus captured by the sampled $K_0$ . However, their performance deteriorates on other tasks, highlighting the limitations of relying on Wikipedia-sampled $K_0$. Our method, which directly incorporates the gated mechanism, offers a more precise and effective approach compared to approximations derived from Wikipedia. For results on more models, please see Appendix \ref{app:more_exper}.

\subsection{Scaling up the number of edited facts}
We further examine the scalability of the NeuralDB when applied to an extensive volume of knowledge. To obtain a sufficiently large set of facts for this investigation, we utilized the training set of the ZsRE dataset for model editing. The results for the Llama3 8B (Instruct) model are presented in Table~\ref{tab:extremely long}. These results demonstrate that the performance of NeuralDB remains highly stable as the number of edited facts increases from 10,000 to 100,000 with only marginal degradation observed. In evaluations of the model’s general ability, we find that scaling up the number of edited facts to 100,000 did not harm the general ability performance and led to a 0.7\% improvement in the benchmark datasets. This underscores the superior scalability of our framework.

\begin{table}[h]
\vspace{-1em}
\caption{Editing accuracy and the post-edited model’s general performance of NeuralDB on Llama 3 (8B) when editing extremely large sets of facts.}
\vskip 0.04in
   \label{tab:extremely long}
\centering
    \begin{tabular}{ccccccccccccc}
    \toprule
        Number of edits& 0k & 10k & 20k & 30k & 40k & 50k &60k&70k&80k&90k&100k \\ \midrule 
        Efficacy $(\uparrow)$&37.0  & 96.9  & 96.6  & 96.6  & 96.4  & 96.1 &96.0 &	95.9 &	95.8 &	95.6  & 95.5  \\ 
        Generalization $(\uparrow)$ & 36.3 & 91.4  & 91.4  & 91.2  & 91.0  & 90.7& 90.6& 	90.6 &	90.5 &	90.4 & 90.2   \\ 
        Specificity $(\uparrow)$ &31.9  & 35.1  & 35.3  & 35.2  & 35.2  & 35.2& 35.2 &	35.2 	&35.1 &	35.1  & 35.1  \\ 
        MMLU\tablefootnote{The MMLU results are evaluated by the AlphaEdit project rather than the lm-evaluation-harness. Although they use different metrics, both sets of results reflect the general capabilities of the LLMs.} $(\uparrow)$&	56.2 &	56.2 &	56.2 &	56.2& 	56.2& 	56.2& 	56.2& 	56.9& 	56.9&	56.9 & 56.9 \\
        \bottomrule
    \end{tabular}

\end{table}

\section{Related Work}

\subsection{Knowledge editing through parameter modification}
\paragraph{Locate-and-edit} The L\&E paradigm is derived from casual trace experiments~\citep{meng2022locating}, suggesting that the factual memory of the Transformer models is primarily associated with the FFN layers~\citep{geva-etal-2021-transformer}, therefore Transformers can be understood using associative memory \citep{niu2024beyond}. ROME~\citep{meng2022locating} is proposed to edit the factual memory of the models by modifying the parameter of the target FFN layer. MEMIT~\citep{MEMIT} extends ROME to support multi-layer and batch editing versions, allowing the editing of thousands of factual knowledge. To address the challenge of post-edited models losing their general capabilities~\citep{li2024should,hsueh-etal-2024-editing}, several solutions were proposed, including the dumping of sequential editing caches~\citep{huang2024reasons}, null space projection~\citep{fang2025alphaedit}, and regularization of the weights~\citep{gu-etal-2024-model} and singular values~\citep{ma2025perturbationrestrained}.

\paragraph{Hypernetwork} KE \citep{de2021editing} trains a lightweight biLSTM-MLP editor to convert a single-sample gradient into a low-rank weight delta. MEND \citep{mitchell2021fast} factorizes two-layer gradients into rank-1 vectors and feeds them through a shared MLP, allowing memory-frugal batch edits. InstructEdit \citep{zhang2024instructedit} prepends '[Task][Description][Input]' prompts so that gradients self-cluster, allowing one low-rank editor to perform reliable updates in diverse OOD tasks. All these approaches require additional fine-tuning of the hypernetwork with large datasets, incurring considerable computational overhead.

\subsection{Knowledge editing without parameter modification}
\paragraph{External module}  This line of work with external memory caches starts with SERAC \citep{mitchell2022memory}, which augments a frozen model with explicit retrieval memory, a scope classifier, and a lightweight counterfactual Seq2Seq generator so that edits are applied by lookup rather than gradient updates. T-Patcher \citep{huang2023transformer} injects a sparse “key–value–bias” triplet per error into the final FFN layer, ensuring each patch fires only on its intended trigger.  GRACE \citep{hartvigsen2023aging} stores erroneous hidden states as discrete keys in a dynamic codebook whose values overwrite selected Transformer layers whenever the current activation falls inside an $\epsilon$ -ball, enabling millisecond-scale, thousand-edit hot fixing with $< 0.4 \%$ extra parameters. MELO \citep{yu2024melo} similarly uses a hidden‐state–indexed database to activate low‐rank, per‐edit adapter blocks on demand. MindBridge \citep{li2025mindbridge} encodes each edit as a standalone “memory modality” that can be shared across future model versions, preventing knowledge loss when the base LLM is upgraded. Similarly, the external modules of these methods require a large number of datasets for fine-tuning, which leads to high costs.

\paragraph{Prompt-based approaches}  
Recent studies utilize prompt engineering to facilitate efficient KE. For example, MemPrompt~\citep{madaan-etal-2022-memory} and IKE~\citep{zheng2023can} embed updated knowledge into prompts to leverage in-context learning. For multi-hop QA tasks, MQUAKE~\citep{zhong2023mquake} introduces a benchmark to evaluate KE performance.
MeLLo~\citep{zhong2023mquake} stores edited facts externally and iteratively prompts the model to yield answers consistent with the updates. PokeMQA~\citep{gu-etal-2024-pokemqa} improves retrieval and answer accuracy by decomposing multi-hop questions via prompts. 
RAE~\citep{10.1145/3627673.3679722} retrieves refined facts and enhances the language model through in-context learning using a knowledge graph. 
To address multilingual KE, ReMaKE~\citep{wang-etal-2024-retrieval} integrates newly retrieved multilingual knowledge into prompts for better adaptation.  


\section{Conclusion}


In this paper, we introduce NeuralDB, a scalable knowledge editing framework designed to construct a neural KV database from edited facts and integrate it into the target FFN layer within LLMs using a non-linear gated function. This integration ensures that the general capabilities of LLMs are preserved. The neural database is designed to be easily maintained, facilitating efficient addition and modification of edited facts within the models. We conducted comprehensive experiments across various LLMs to validate the effectiveness of our framework. Our results on the ZsRE and CounterFacts datasets, utilizing GPT2-XL, GPT-J (6B), and Llama-3 (8B), demonstrate that NeuralDB editing can effectively modify hundreds of thousands of facts without degrading the quality of generated text. Additionally, our findings from six generic text understanding and generation tasks confirm that our method preserves the general abilities of LLMs unrelated to the target edited facts. These results highlight the robustness and scalability of NeuralDB editing, positioning it as a valuable tool for enhancing the adaptability and accuracy of LLMs in diverse applications.
\bibliography{ref.bib}
\bibliographystyle{unsrtnat}

\newpage
\appendix

\section{Limitation}\label{App:limitation}
Although NeuralDB can effectively edit thousands of pieces of knowledge, the corresponding storage overhead grows linearly with the number of edited entries. This may lead to additional memory usage, especially when editing a large amount of knowledge. For instance, after editing 100,000 pieces of knowledge for LLama3 8B, the additional memory usage reached 20\% of the original model.

\section{Border impact}\label{App:border impact}

As information evolves rapidly in society, LLMs must also adapt swiftly. Pre-training a new LLM requires a substantial amount of time and resources. Knowledge editing enables models to update information and enhance their domain-specific knowledge. Our approach significantly extends the upper limit of the number of edits that can be made without compromising the model's performance. This advancement has the potential to drive broader applications of knowledge editing.

The wider impacts include positive contributions to society. By improving the efficiency of knowledge updates, models can adapt more quickly to changing environments and information. This not only benefits research and education, but also provides up-to-date information support in critical fields such as healthcare and justice, promoting scientific and timely decision making. Furthermore, the proliferation of knowledge editing will encourage interdisciplinary collaboration, allowing experts from different fields to share and integrate knowledge more effectively to address complex societal issues.

As the capabilities for knowledge editing expand, it is imperative to address the potential ethical and social implications. Ensuring the accuracy and credibility of information is essential to prevent the dissemination of misinformation. Furthermore, a transparent editing process and traceability are crucial for building public trust and ensuring responsible use of technology. The proposed NeuralDB editing framework, explicitly derived from updated knowledge, facilitates interpretable editing. Consequently, our research not only focuses on technological advancement but also emphasizes the broader applications and societal impacts of these innovations.

\section{Baselines}\label{App:baselines}
In this section, we present the six baseline methods evaluated in our work. For each method, we adopt the default hyperparameter settings provided in the official code of the corresponding papers.
\begin{itemize}
\item \textbf{Fine-Tune (FT)}~\citep{MEMIT} is a fine-tuning method that updates the FFN of a transformer layer to incorporate new factual knowledge. The target layer is selected on the basis of its relevance to the knowledge being edited. FT operates by maximizing the likelihood of the target output using the standard next-token prediction loss.
\item \textbf{MEND}~\citep{MEND} introduces a hypernetwork that maps fine-tuning gradients into efficient weight updates for a pre-trained model. By applying low-rank decomposition to the gradients, it reduces parameter complexity and enables lightweight, localized edits without full model retraining.
\item \textbf{ROME}~\citep{meng2022locating} performs KE by interpreting the FFN in a transformer layer as a linear associative memory. It derives key-value pairs from internal activations and computes a weight update that ensures the edited layer produces the desired hidden representation. A rank-one modification is then applied to the FFN weights, aligning the model’s internal representations with the new factual knowledge.
\item \textbf{MEMIT}~\citep{MEMIT} extends the ROME framework to support simultaneous editing of a large number of factual knowledge items. It models the updates as a joint optimization over key-value pairs and applies rank-one modifications to the FFNs. To prevent interference between edits, MEMIT distributes the updates in a top-down manner across critical FFN layers, achieving efficient, scalable, and stable insertion of new factual knowledge. 
\item \textbf{RECT}~\citep{gu2025editing} reduces the degradation of general capabilities caused by KE. It regularizes weight updates by constraining their magnitude and selectively updates only the top-$k$\% of parameters with the largest changes during fine-tuning. This reduces overfitting and helps preserve the model’s reasoning and question-answering abilities, while still achieving effective factual edits.
\item \textbf{AlphaEdit}~\citep{fang2025alphaedit} introduces a null-space projection mechanism to preserve existing knowledge during editing. It projects the update direction onto the null space of prior knowledge and then applies the projected perturbation to model parameters. This approach reduces interference with previously edited facts and enables effective integration of new information. 
\end{itemize}

\section{Datasets and Metrics}\label{App:datasets and metrics}
In this section, we describe the datasets and evaluation metrics employed in our experiments.
\subsection{Datasets}
We evaluate our methods on two types of datasets: CounterFact and ZsRE for assessing KE, and six benchmarks including SciQ, MMLU, CommonsenseQA, ARC Challenge, WSC273, and LAMBADA for evaluating the general capabilities of post-edited models.
\begin{itemize}
\item \textbf{Counterfact}~\citep{meng2022locating} is a challenging benchmark that focuses on editing incorrect factual statements in language models. Each instance includes a subject and an incorrect attribute to be updated. To assess edit locality, it provides contrastive prompts involving related but distinct entities, ensuring changes do not affect nearby facts. Additionally, the dataset includes paraphrased and semantically equivalent prompts to evaluate the generalization, fluency, and consistency of the edited model responses.
\item \textbf{ZsRE}~\citep{zsre} is a question-answering dataset commonly used in knowledge editing evaluation. Each example includes a subject, a target answer to be edited, paraphrased questions for testing generalization, and unrelated questions for evaluating locality. The dataset also features human-written question variants, enabling assessment of model robustness to semantically equivalent inputs.

\item \textbf{SciQ}~\citep{welbl-etal-2017-crowdsourcing} is a multiple-choice science QA dataset covering topics such as physics, chemistry, and biology. It is used to evaluate a model’s ability to recall factual scientific knowledge.

\item \textbf{MMLU}~\citep{hendrycks2021measuring} is a multitask benchmark containing questions from 57 academic and professional disciplines. It assesses factual knowledge and generalization across diverse domains in zero- and few-shot settings.

\item \textbf{CommonsenseQA}~\citep{talmor-etal-2019-commonsenseqa} is a multiple-choice QA dataset that evaluates a model’s ability to reason over everyday commonsense. It focuses on applying implicit world knowledge to select the correct answer among distractors.

\item \textbf{ARC Challenge}~\citep{clark2018think} is a science QA dataset designed to require reasoning beyond simple retrieval. It contains complex grade-school level questions that challenge a model’s problem-solving abilities.

\item \textbf{WSC273}~\citep{kocijan-etal-2019-surprisingly} is a coreference resolution benchmark derived from the Winograd Schema Challenge. It is designed to test whether a model can correctly identify what ambiguous pronouns refer to, based on context and commonsense reasoning.

\item \textbf{Lambada}~\citep{paperno-etal-2016-lambada} is a word prediction benchmark composed of narrative passages where the final word can only be inferred from the entire context. It is designed to evaluate a model’s ability to capture long-range dependencies and maintain discourse-level coherence.
\end{itemize}
\subsection{Metrics}
Here, we introduce the evaluation metrics for the CounterFact and ZsRE datasets, which are selected based on previous works~\citep{meng2022locating,fang2025alphaedit}.

\subsubsection{CounterFact Metrics}
Given a language model \( f \), a query \((s_i, r_i)\), an edited object \(\hat{o}_i\), and the original object \(o_i\), the evaluation metrics for CounterFact are defined as follows.
\begin{itemize}
    \item \textbf{Efficacy (success of edited facts):}  
    The proportion of instances in which the model prefers the edited object \(\hat{o}_i\) over the original object \(o_i\) when prompted with \((s_i, r_i)\):
    \begin{equation}
        \mathbb{E}_i \left[ \mathbb{P}_{f}\left[\hat{o}_i \mid (s_i, r_i)\right] > \mathbb{P}_f\left[o_i \mid (s_i, r_i)\right] \right].
    \end{equation}

    \item \textbf{Generalization (success of paraphrased facts):}  
    The proportion of paraphrased prompts \(F_i(s_i, r_i)\), representing rephrasings of the original query \((s_i, r_i)\), for which the model assigns higher likelihood to \(\hat{o}_i\) than to \(o_i\):
    \begin{equation}
        \mathbb{E}_i \left[ \mathbb{P}_{f}\left[\hat{o}_i \mid F_i(s_i, r_i)\right] > \mathbb{P}_{f}\left[o_i \mid F_i(s_i, r_i)\right] \right].
    \end{equation}

    \item \textbf{Specificity (success of neighborhood facts):}  
    The proportion of neighborhood prompts \(N_i(s_i, r_i)\), which involve subjects semantically related to but distinct from the original subject \(s_i\), for which the model assigns higher likelihood to the correct object \(o_i\) than to the edited object \(\hat{o}_i\):
    \begin{equation}
        \mathbb{E}_i \left[ \mathbb{P}_{f}\left[\hat{o}_i \mid N_i(s_i, r_i)\right] < \mathbb{P}_{f}\left[o_i \mid N_i(s_i, r_i)\right] \right].
    \end{equation}

    \item \textbf{Fluency (generation entropy):}  
    Measures output repetition based on the entropy of n-gram distributions in model outputs. Specifically, it computes a weighted combination of bigram and trigram entropies, where \(g_n(\cdot)\) denotes the n-gram frequency distribution:
    \begin{equation}
        -\frac{2}{3} \sum_{k} g_2(k) \log_2 g_2(k) + \frac{4}{3} \sum_{k} g_3(k) \log_2 g_3(k).
    \end{equation}

    \item \textbf{Consistency (reference score):}  
    Consistency is measured by prompting the model \(f\) with a subject \(s\) and computing the cosine similarity between the TF-IDF vectors of its generated output and a reference Wikipedia passage about the object \(o\).
\end{itemize}

\subsubsection{ZsRE Metrics}
Given a language model \( f \), a query \((s_i, r_i)\), an edited object \(\hat{o}_i\), and the original object \(o_i\), the evaluation metrics for ZsRE are defined as follows:
\begin{itemize}
    \item \textbf{Efficacy (success of edited facts):}  
    Top-1 accuracy on the edited samples, measuring the proportion of cases in which the model ranks the edited object \(\hat{o}_i\) as the most likely prediction given the prompt \((s_i, r_i)\):
    \begin{equation}
        \mathbb{E}_i \left[ \hat{o}_i = \arg\max_o \mathbb{P}_f\left(o \mid (s_i, r_i)\right) \right]
    \end{equation}

    \item \textbf{Generalization (success of paraphrased facts):}  
    Top-1 accuracy on paraphrased prompts \(F_i(s_i, r_i)\), which are rephrasings of the original query \((s_i, r_i)\), measuring the proportion of cases in which the model ranks the edited object \(\hat{o}_i\) as the most likely prediction for the given rephrased prompt:
    \begin{equation}
        \mathbb{E}_i \left[ \hat{o}_i = \arg\max_o \mathbb{P}_f\left(o \mid F_i(s_i, r_i)\right) \right]
    \end{equation}

    \item \textbf{Specificity (success of neighborhood facts):}  
    Top-1 accuracy on neighborhood prompts \(N_i(s_i, r_i)\), which involve subjects related to but distinct from \(s_i\). Specificity reflects whether the model preserves correct predictions on unaffected inputs by still preferring \(o_i\) over \(\hat{o}_i\):
    \begin{equation}
        \mathbb{E}_i \left[ o_i = \arg\max_o \mathbb{P}_f\left(o \mid N_i(s_i, r_i)\right) \right]
    \end{equation}
\end{itemize}

\section{Implementation details}\label{App:implementation details}
We provide the details of the implementation of the experiments. To reproduce our methods, a GPU with 40G memory is required. Our framework is built on L\&E methods like MEMIT~\citep{MEMIT} and AlphaEdit~\citep{fang2025alphaedit}. We first sequentially compute the key vector $\mathbf{k}$ and the residual vector $\mathbf{r}$ for the target edited facts. The details of the computation can be found in Appendix~\ref{App:kvr compute}. Then we stack them as the key matrix $\mathbf{K}_1$ and the residual matrix $\mathbf{R}_1$ and construct a neural KV database $(\mathbf{K}_1, \mathbf{R}_1)$. Then we integrate the non-linear retrieval module with the target FFN layer $l^*$. Our module only involves one hyperparameter $\gamma$ to control the gated mechanisms. We provide the details of the setting in the following Table~\ref{tab:hyper-para}.

\begin{table}[h]
    \centering
\caption{Hyper-parameters of NeuralDB for various models in the main experiments}\label{tab:hyper-para}
\vskip 0.1in
\begin{tabular}{cccc}
\toprule
                              & GPT2-xl & GPT-J (6B) & Llama 3 Instruct (8B) \\ \midrule
Layer found by casual trace & 17      & 17         & 17                    \\  
Layer $l^*$ used by NeuralDB                          & 17      & 8          & 7                     \\ 
$\gamma$                      & 0.65    & 0.65       & 0.65           \\
\bottomrule
\end{tabular}
\end{table}

\section{The computation of $\mathbf{k}_i$ and $\mathbf{r}_i$}\label{App:kvr compute}
We follow previous locating-and-editing methods~\citep{meng2022locating,MEMIT,fang2025alphaedit} to derive the key vector and residual vector from the given edited fact $(s_i,r_i,o_i \rightarrow  \hat{o}_i)$. Let ${l^*}$ denote the FFN layer to be updated. 

For the key vector $\mathbf{k}_i$, we retrieve the specified activation from LLM inferring the prompt.  We denote $\mathbf{k}^{l^*}(\mathbf{x})$ as the key activation of the prompt $\mathbf{x}$ in layer ${l^*}$. Then the target key vector are computed by the following average over random prefix $\mathbf{x}_j$:
\begin{equation}
    \mathbf{k}_i = \frac{1}{N} \sum_{j=1}^N \mathbf{k}^{l^*}(\mathbf{x}_j+ \mathbf{s}_i),
\end{equation}
where $\mathbf{s}_i$ is the subject of edited fact. The $\mathbf{x}_j$ is the prefix randomly generated by the language model $f$ to improve the robustness of the expressive ability of $\mathbf{k}_i$.

For the target vector $\mathbf{r}_i$, we wish to find some vector to decode the new answer $\hat{o}_i$. We utilize the supervised learning to  derive  $\mathbf{r}_i = \arg \min_{\mathbf{r}} L(\mathbf{r})$, where the loss object $L(\mathbf{r})$ is defined as following:
\begin{equation}
     \frac{1}{N} \sum_{j=1}^N \left(-\text{log} ~\mathbb{P}_{f(\mathbf{h}^{l^{*}} += \mathbf{r})} [o^*|x_j+p] + D_{\text{KL}}( \mathbb{P}_{f(\mathbf{h}^{l^{*}} += \mathbf{r})} [x|p']\|\mathbb{P}_f [x|p'])\right).
\end{equation}
$p$ is the factual prompt while $p'$ is its variant ( the form of ``{subject} is a''). $f(\mathbf{h}^{l^{*}} += \mathbf{r})$ indicates substituting the output of the $i$-th MLP with an additional learnable parameter $\mathbf{r}$. This optimization also uses the random prefix text $x_j$ to enhance the robustness.
\section{Additional memory usage and computation time}~\label{App: addional memory and time}
In this section, we provide a detailed discussion of additional resources of our new module.

\paragraph{Memory usage} We cache the key matrix $\mathbf{K}_1$ and the residual matrix $\mathbf{R}_1$ and construct the new module, which totally take $(d_1+d_2) \times m$ parameters with $m$ denoting the number of edited facts. For Llama 3 8B model with $d_1 = 14,336$, $d_2 = 4,096$, the memory of 10,000 facts is about 150 million parameters. Compared to the total 8B parameters, the additional memory for 1M facts is only $2.2\%$. Additionally, our memory grows linearly with the facts and is easily scaled to more facts.

\paragraph{Computation time} We report the average evaluation time for three models and two datasets in Table~\ref{tab:addtion time}. The results show that the averaged time has only a slight improvement compared with the methods without an additional module.
\begin{table}[h]
    \centering
    \caption{The averaged time of evaluation post-edited models on CounterFacts and ZsRE}~\label{tab:addtion time}
    \vskip 0.1in
    \begin{tabular}{c|ccc|ccc}
    \toprule
        Model & \multicolumn{3}{c|}{Llama3}  &  \multicolumn{3}{c}{GPTJ-6B}  \\ \midrule
        Method & MEMIT & AlphaEdit & NeuralDB & MEMIT & AlphaEdit & NeuralDB \\ \midrule
        CounterFacts & 4.12 & 4.11 & 4.18 & 3.81 & 3.76 & 3.90 \\ 
        ZsRE & 0.22 & 0.22 & 0.22 & 0.16 & 0.16 & 0.17 \\ \bottomrule
    \end{tabular}
\end{table}
\section{Ablation study}~\label{app:abalation study}
\subsection{The impact of layer selection and the setting of the parameter $\gamma$}

\begin{table}[h]
    \centering
    \caption{Ablation study on Llama 3 (8B)}
    \label{tab:ablation_study}
    \begin{tabular}{ccccccc}
    \toprule
        Gamma & Layer & E  & G & S& F& C  \\ \midrule
        0.65 & 7 & 99.2 & 85.9 & 85.6 & 631.9 & 32.6  \\ 
        0.65 & 8 & 99.2 & 79.3 & 85.1 & 631.5 & 33.3  \\ 
        0.65 & 9 & 99.2 & 77.4 & 84.9 & 631.6 & 32.4  \\ 
        0.55 & 7 & 99.1 & 91.9 & 83.4 & 631.6 & 32.7  \\ 
        0.75 & 7 & 99.2 & 74.1 & 86.2 & 632.3 & 32.2 \\ \bottomrule
    \end{tabular}
\end{table}
We conducted an ablation study to investigate the choice of hyperparameters, including $\gamma$ and the target layer, with results provided in Table~\ref{tab:ablation_study}. For the configuration shown in Table~\ref{tab:overall_comp}, where $\gamma = 0.65$ and $L=7$ for Llama3, we vary $\gamma$ between 0.55 and 0.75, and layer between 8 and 9. Although layer 8 is determined by the causal trace, our results show that its results are not suboptimal. An increase in $\gamma$ can improve the specificity while damaging the generation, which aligns the gated mechanism. In general, the results demonstrate that it is necessary to search for optimal hyperparameters.
\subsection{Multi-layer editing}
\begin{algorithm}[t]
    \caption{Old multi-layer method}
    \begin{algorithmic}[1]  
        \REQUIRE Input Transformer model $f$, target layers list $L=[l_1, \cdots, l_n]$, request facts $\mathcal{F}$, Function \textsc{Compute\_key} to compute the keys of facts at layer $l$, \textsc{Compute\_residual} compute the residual of facts at layer $l$ .
        \STATE $R \gets \textsc{Compute\_residual}(f, \mathcal{F}, l_n )$

        \FOR{$l$ in $L$}
            \STATE $K_i \gets \textsc{Compute\_key}(f, \mathcal{F}, l )$
            \STATE $R_i \gets  \text{R} / (l_n - l + 1)$
            \STATE Perform KE at layer $l$ with $(K_i,R_i)$ \\
        \ENDFOR
    \end{algorithmic}
\end{algorithm}

\begin{algorithm}[t]
    \caption{New multi-layer method}
    \begin{algorithmic}[1]  
        \REQUIRE Input Transformer model $f$, target layers list $L=[l_1, \cdots, l_n]$, request facts $\mathcal{F}$, Function \textsc{Compute\_key} to compute the keys of facts at layer $l$, \textsc{Compute\_residual} compute the residual of facts at layer $l$ .

        \FOR{$l$ in $L$}
            \STATE $K_i \gets \textsc{Compute\_key}(f, \mathcal{F}, l )$
        \STATE $R_i \gets \textsc{Compute\_residual}(f, \mathcal{F}, l )$
            \STATE Perform KE at layer $l$ with $(K_i,R_i)$ \\
        \ENDFOR
    \end{algorithmic}
\end{algorithm}

\begin{table}[ht]
  \centering
  \caption{Editing performance under different layer setup}
  \vskip 0.1in
  \label{tab:edit-comparison}
  \begin{tabular}{llcccc}
    \toprule
    Model    & Layer Setup               & Efficacy ↑ & Generalization ↑ & Specificity ↑ & Fluency ↑ \\ 
    \midrule
    \multirow{3}{*}{GPT-J}%
             & [8] baseline               & 99.08      & 93.48            & 75.52         & 620.53    \\
             & [6,7,8] new multi layers   & 94.44      & 91.72            & 75.93         & 617.44    \\
             & [6,7,8] old multi layers   & 99.31      & 93.23            & 76.78         & 616.00    \\
    \midrule
    \multirow{3}{*}{GPT2-XL}%
             & [17] baseline              & 99.04      & 95.96            & 70.72         & 621.90    \\
             & [15,16,17] new multi layers& 94.81      & 92.68            & 70.26         & 618.51    \\
             & [15,16,17] old multi layers& 99.08      & 94.01            & 71.33         & 624.48    \\
    \bottomrule
  \end{tabular}
\end{table}

\begin{table}[t]
\centering
\scriptsize
\caption{Performance of each task across different editing budgets (1,000, 2,000, 4,000, 6,000, 8,000, 10,000) under various model–algorithm configurations.}
\label{tab:subtables}

\begin{subtable}[t]{\textwidth}
  \centering
  \caption{GPT-J (AlphaEdit)}
  \begin{tabular}{l*{6}{S}}
    \toprule
    {Task} 
      & {1,000}    & {2,000}    & {4,000}    & {6,000}    & {8,000}    & {10,000}   \\
    \midrule
    sciq            & 0.9110 & 0.9080 & 0.9060 & 0.8900 & 0.8850 & 0.7410 \\
    logiq\_a        & 0.2151 & 0.2243 & 0.2089 & 0.2181 & 0.2304 & 0.2181 \\
    commonsense\_qa & 0.2080 & 0.2146 & 0.2048 & 0.1884 & 0.1925 & 0.1785 \\
    arc\_easy       & 0.6658 & 0.6477 & 0.6326 & 0.6010 & 0.5804 & 0.4870 \\
    MMLU            & 0.2660 & 0.2688 & 0.2622 & 0.2592 & 0.2587 & 0.2535 \\
    arc\_challenge  & 0.3276 & 0.3148 & 0.2901 & 0.2782 & 0.2611 & 0.2261 \\
    lambada         & 0.6722 & 0.6604 & 0.6057 & 0.5158 & 0.4036 & 0.2203 \\
    winogrande      & 0.6346 & 0.6227 & 0.6093 & 0.5991 & 0.5730 & 0.5635 \\
    wsc273          & 0.8425 & 0.8352 & 0.7985 & 0.7399 & 0.7179 & 0.6264 \\
    \bottomrule
  \end{tabular}
\end{subtable}

\vspace{1em}

\begin{subtable}[t]{\textwidth}
  \centering
  \caption{GPT-J (NeuralDB)}
  \begin{tabular}{l*{6}{S}}
    \toprule
    {Task} 
      & {1,000}    & {2,000}    & {4,000}    & {6,000}    & {8,000}    & {10,000}   \\
    \midrule
    sciq            & 0.9160 & 0.9160 & 0.9160 & 0.9160 & 0.9160 & 0.9160 \\
    logiq\_a        & 0.2120 & 0.2120 & 0.2120 & 0.2120 & 0.2120 & 0.2120 \\
    commonsense\_qa & 0.2080 & 0.2080 & 0.2080 & 0.2080 & 0.2080 & 0.2080 \\
    arc\_easy       & 0.6692 & 0.6692 & 0.6692 & 0.6692 & 0.6692 & 0.6692 \\
    MMLU            & 0.2695 & 0.2697 & 0.2697 & 0.2695 & 0.2695 & 0.2698 \\
    arc\_challenge  & 0.3396 & 0.3396 & 0.3404 & 0.3404 & 0.3404 & 0.3404 \\
    lambada         & 0.6829 & 0.6827 & 0.6827 & 0.6821 & 0.6821 & 0.6819 \\
    winogrande      & 0.6409 & 0.6417 & 0.6417 & 0.6417 & 0.6417 & 0.6409 \\
    wsc273          & 0.8242 & 0.8242 & 0.8242 & 0.8242 & 0.8242 & 0.8242 \\
    \bottomrule
  \end{tabular}
\end{subtable}

\vspace{1em}

\begin{subtable}[t]{\textwidth}
  \centering
  \caption{GPT-2 XL (AlphaEdit)}
  \begin{tabular}{l*{6}{S}}
    \toprule
    {Task} 
      & {1,000}    & {2,000}    & {4,000}    & {6,000}    & {8,000}    & {10,000}   \\
    \midrule
    sciq            & 0.8250 & 0.8230 & 0.7920 & 0.7440 & 0.6390 & 0.4920 \\
    logiq\_a        & 0.2289 & 0.2273 & 0.2012 & 0.2104 & 0.1951 & 0.1889 \\
    commonsense\_qa & 0.1908 & 0.1957 & 0.1916 & 0.1974 & 0.2080 & 0.1933 \\
    arc\_easy       & 0.5682 & 0.5484 & 0.4987 & 0.4693 & 0.4066 & 0.3493 \\
    MMLU            & 0.2618 & 0.2562 & 0.2464 & 0.2312 & 0.2369 & 0.2315 \\
    arc\_challenge  & 0.2423 & 0.2346 & 0.2398 & 0.2108 & 0.1887 & 0.2065 \\
    lambada         & 0.4881 & 0.4170 & 0.2610 & 0.1467 & 0.0767 & 0.0231 \\
    winogrande      & 0.5904 & 0.5549 & 0.5564 & 0.5272 & 0.5201 & 0.5067 \\
    wsc273          & 0.6520 & 0.6227 & 0.5714 & 0.5861 & 0.5678 & 0.5421 \\
    \bottomrule
  \end{tabular}
\end{subtable}

\vspace{1em}

\begin{subtable}[t]{\textwidth}
  \centering
  \caption{GPT-2 XL (NeuralDB)}
  \begin{tabular}{l*{6}{S}}
    \toprule
    {Task} 
      & {1,000}    & {2,000}    & {4,000}    & {6,000}    & {8,000}    & {10,000}   \\
    \midrule
    sciq            & 0.8240 & 0.8290 & 0.8280 & 0.8280 & 0.8280 & 0.8280 \\
    logiq\_a        & 0.2212 & 0.2181 & 0.2181 & 0.2181 & 0.2181 & 0.2197 \\
    commonsense\_qa & 0.1900 & 0.1933 & 0.1941 & 0.1941 & 0.1941 & 0.1941 \\
    arc\_easy       & 0.5770 & 0.5785 & 0.5848 & 0.5848 & 0.5848 & 0.5848 \\
    MMLU            & 0.2532 & 0.2545 & 0.2547 & 0.2546 & 0.2543 & 0.2544 \\
    arc\_challenge  & 0.2509 & 0.2509 & 0.2491 & 0.2500 & 0.2500 & 0.2517 \\
    lambada         & 0.5055 & 0.5053 & 0.5077 & 0.5069 & 0.5065 & 0.5053 \\
    winogrande      & 0.5770 & 0.5785 & 0.5848 & 0.5848 & 0.5848 & 0.5848 \\
    wsc273          & 0.6777 & 0.6667 & 0.6850 & 0.6850 & 0.6850 & 0.6850 \\
    \bottomrule
  \end{tabular}
\end{subtable}

\end{table}

We also propose two algorithms for multi-layer updating, and compare these two algorithms on two different pre-trained models in Table~\ref{tab:edit-comparison}. The experimental results indicate that the new multi-layer method can greatly scale the number of editable facts, albeit with some loss in performance, while the old multi-layer method—though achieving better editing accuracy—requires substantially more storage.

\section{Additional experiments}~\label{app:more_exper}

\subsection{lm-evaluation-hardness}
We conducted more experiments on lm-evaluation-hardness using GPT2-XL and GPT-J, see Table \ref{tab:subtables} for details.

\subsection{Weighted score visualization of paraphrased and neighborhood facts}\label{subsec: weight score}

\begin{figure}[t]
    \centering
    \begin{minipage}[b]{\linewidth}
        \centering
        \includegraphics[width=0.9\linewidth]{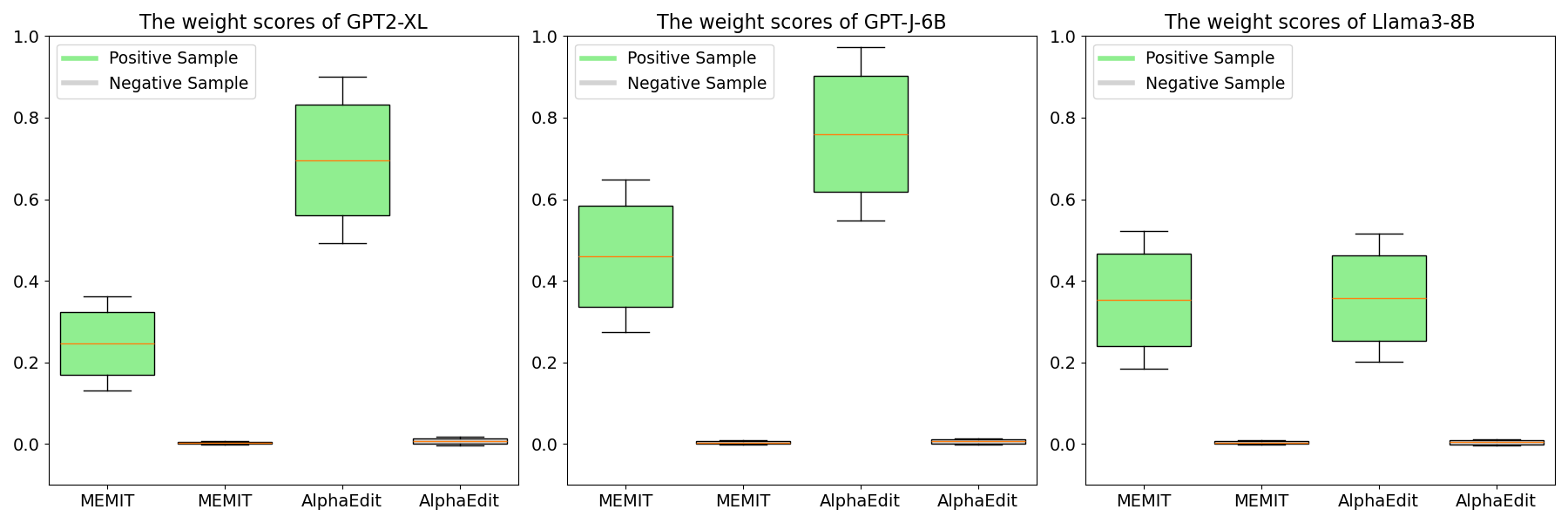}
        \subcaption{Paraphrased facts}
        \vspace{1em}
    \end{minipage}
    
    \begin{minipage}[b]{\linewidth}
        \centering
        \includegraphics[width=0.9\linewidth]{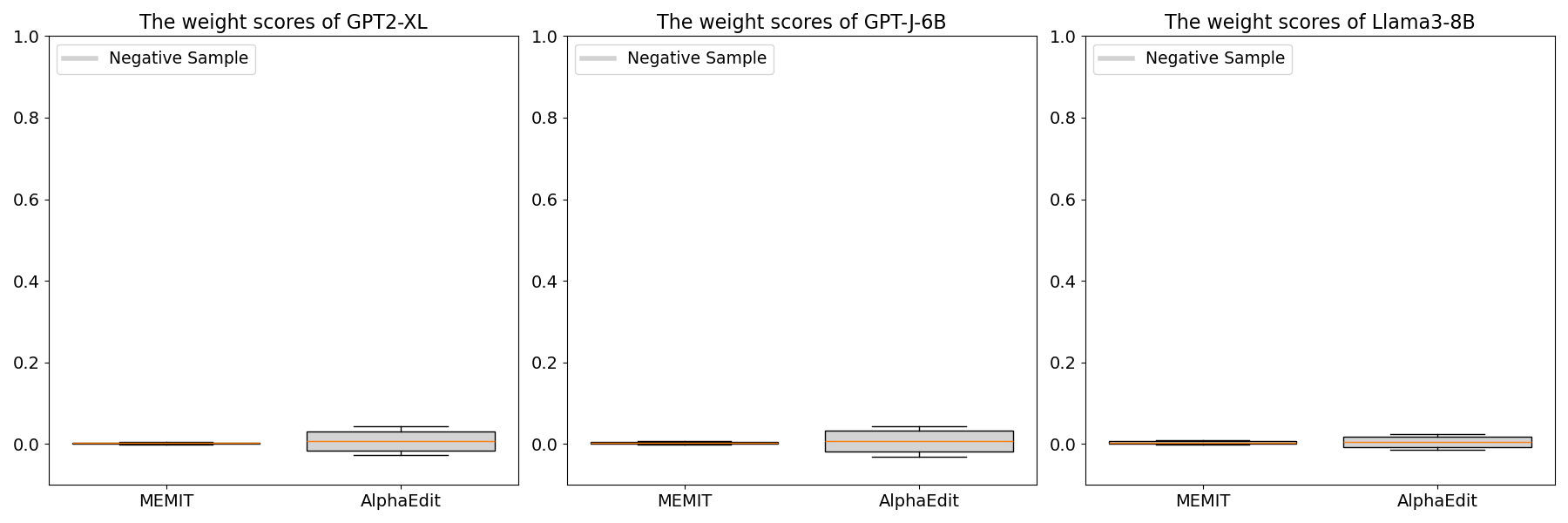}
        \subcaption{Neighborhood facts}
    \end{minipage}
    
    \caption{Visualization of weighted scores for paraphrased facts and neighborhood facts, using MEMIT and AlphaEdit across three models. The boxplots are generated from the mean and variance of weight scores, with the center line indicating the mean, boxes showing ±1 standard deviation, and whiskers ±1.5.}
    \label{fig:fact-type-comparison}
    \vspace{-1em}
\end{figure}

We further conduct experiments on paraphrased and neighborhood facts to examine the distribution of weighted scores under both MEMIT and AlphaEdit. As shown in Fig.~\ref{fig:fact-type-comparison}, the scores for positive samples in paraphrased facts are consistently higher, while those for negative samples remain close to $0$. For neighborhood facts, where all components are considered negative, the scores are likewise consistently close to $0$. These results confirm that, during inference, residuals unrelated to the edited facts remain inactive, resulting in near-zero weighted scores.

\setcitestyle{square,numbers,comma}
\end{document}